\documentclass[runningheads]{llncs}
\usepackage[misc]{ifsym}

\usepackage{multirow}
\usepackage{graphicx}
\begin{document}
\title{SEOVER: Sentence-level Emotion Orientation Vector based Conversation Emotion Recognition Model}
\titlerunning{SEOVER}
\author{Zaijing Li\inst{1} \and
Fengxiao Tang\inst{1}\textsuperscript{(\Letter)} \and
Tieyu Sun\inst{3} \and Yusen Zhu\inst{2} \and
Ming Zhao\inst{1}\textsuperscript{(\Letter)}}
\tocauthor{Zaijing~Li, Fengxiao~Tang, Tieyu~Sun, Yusen~Zhu, and Ming~Zhao}
\toctitle{SEOVER: Sentence-level Emotion Orientation Vector based Conversation Emotion Recognition Model}
\authorrunning{Z.Li et al.}
\institute{School of Computer Science, Central South University, Changsha, China \\
	\email{\{lizaijing,tangfengxiao,meanzhao\}@csu.edu.cn}	\and
	School of Mathematics, Hunan University, Changsha, China \\
	\email{zhu\_yusen@163.com}\\
	\and Prevision Technology Limited, Hong Kong, China\\
	\email{scarsty@gmail.com}}
\maketitle              % typeset the header of the contribution
\begin{abstract}
In this paper, we propose a new expression paradigm of sentence-level emotion orientation vector to model the potential correlation of emotions between sentence vectors. Based on it, we design an emotion recognition model referred to as SEOVER, which extracts the sentence-level emotion orientation vectors from the pre-trained language model and jointly learns from the dialogue sentiment analysis model and extracted sentence-level emotion orientation vectors to identify the speaker’s emotional orientation during the conversation. We conduct experiments on two benchmark datasets and compare them with the five baseline models. The experimental results show that our model has better performance on all data sets.

\end{abstract}
\section{Introduction}
Conversation emotion recognition (CER) refers to the process of identifying the speaker's emotions through text, audio, and visual information in the process of two or more persons' conversations. Nowadays, conversation emotion recognition tasks are widely used in social media such as Twitter and Facebook. 

Recent work of conversation emotion recognition mainly focuses on speaker identification and conversation relationship modeling (Majumder et al. 2019; Ghosal et al. 2019). However, these proposals use CNN (Kim,2014) to encode the utterances, which can not express the grammatical and semantic features of the utterance well and lead to inaccurate emotion identification. To mitigate this issue, some works try to employ the BERT (Devlin et al. 2019) model with improved semantic features extraction ability to encode the sentences (Yuzhao Mao et al. 2020; Jiangnan Li et al. 2020). The proposal achieves good results, however, the proposed BERT-based CER does not fully extract the correlation of emotional tendency between sentences especially when the emotion turns in sudden. As shown in Figure 1, we encode the first sentence of the first dialogue of the MELD dataset with CNN and BERT models respectively to obtain a 600-dimensional vector, and normalize the data of each dimension so that its value is a continuous number within (0,1). Then, a hot zone map is leveraged to show the difference between the feature maps encoded with the two models. It is obvious that trends of the feature maps are totally opposite. It seems that those encoding methods and corresponding feature maps lose some important features of the sentences and can't represent the emotional tendency of the utterance well. 

\begin{figure}
	\includegraphics[width=\textwidth]{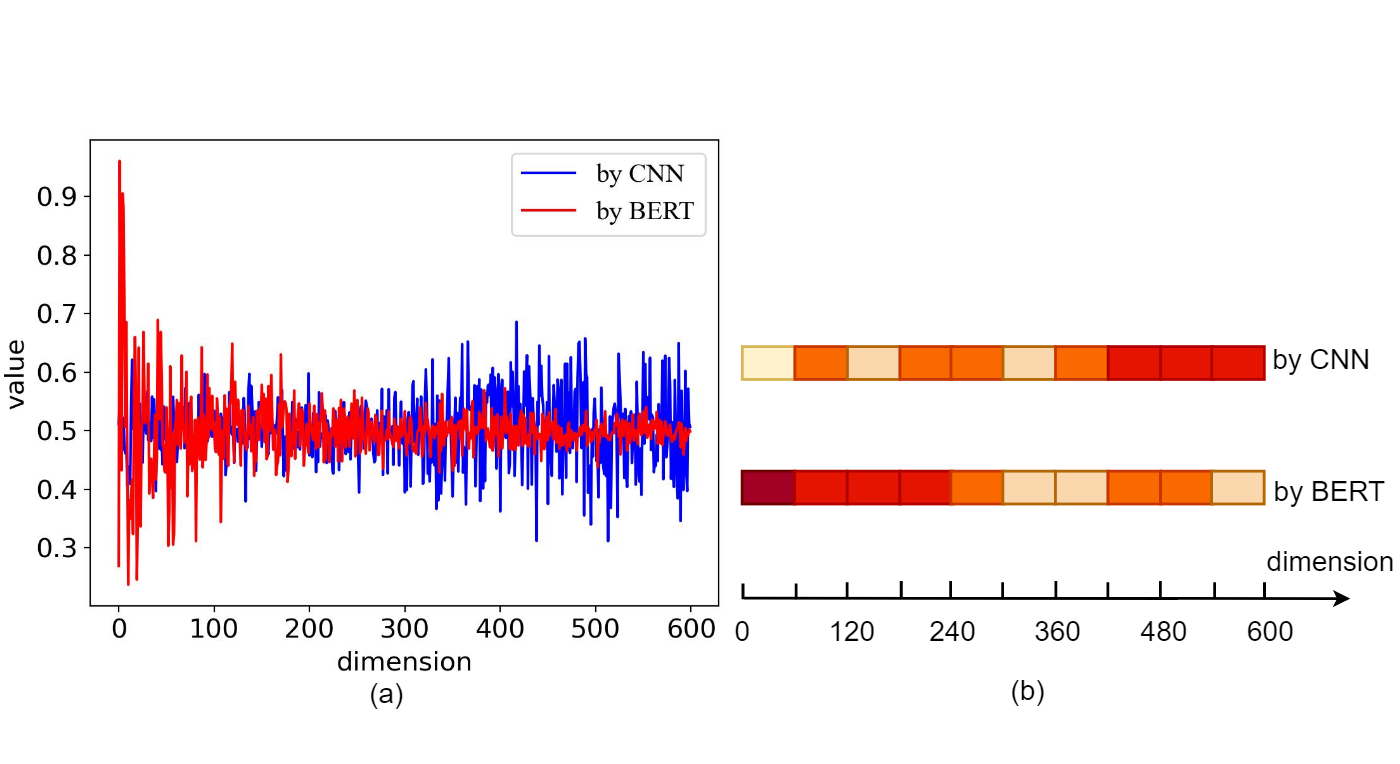}
	\caption{(a) Comparison of the value of each dimension of the sentence vector get by CNN and BERT. (b) Hot zone map of features. The shade of the color indicates the size of the value, the larger the value, the darker the color.} \label{fig1}
\end{figure}
So, we propose a new utterance representation vector referred to as the sentence-level emotion orientation vector (SEOV) to further represent the potential emotion correlation of sentences. The SEOV represents the emotional intensity of sentence encoding vector with its "size", and models the emotion tendency between vectors with its "direction", which is as shown in Figure 2.
\begin{figure}
	\centering
	\includegraphics[width=0.7\textwidth]{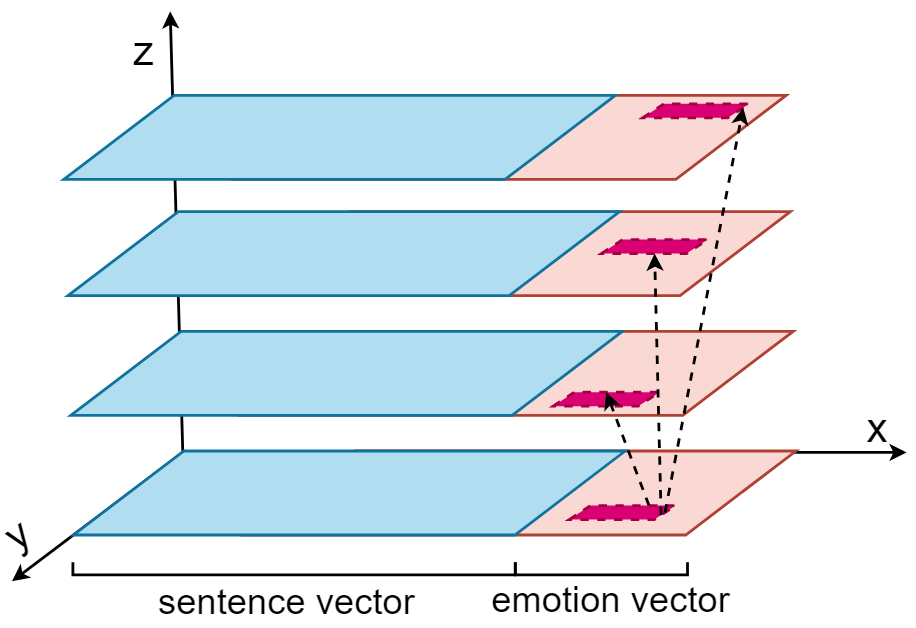}
	\caption{The schematic diagram of SEOV, where the x-axis represents the dimension of the vector, y-axis represents the value of each dimension of the vector, and z-axis represents different sentence-level emotion orientation vectors. There is a correlation between different emotion vectors, indicating the "direction" of the SEOV.} \label{fig2}
\end{figure}

Based on the proposed SEOV, we further propose a sentence-level emotion orientation vector based conversation emotion recognition model (SEOVER) to encode and decode the utterances of dialogue, and get the emotions of each speaker. In the model, we employ an improved transformer called as transformer-emo to extract the SEOV and then jointly use the SEOV and dialogue sentiment analysis model (DSAM) to obtain the contextual semantic information. With continuous fine-tune, we finally get the speaker's emotion classification result.

\section{Methodology}
\subsection{Problem Definition}
Given a conversation $U$: ${U_1, U_2, U_3,...U_n}$, $N$ is the number of conversations, the speaker’s utterance is represented by the function $P_t(U_i)$, where $t$ is the t-th speaker. Our task is to input each utterance $U_i$ and to get its correct classification result in the emotional label set $L$: ${l_1, l_2, l_3, ... l_m}$, where $M$ is the number of types of emotional label.

\subsection{Model}

As shown in the Figure 3, our model is divided into three parts: Sentence-level encoder, emotion-level encoder and context modeling.
\begin{figure}
	\includegraphics[width=\textwidth]{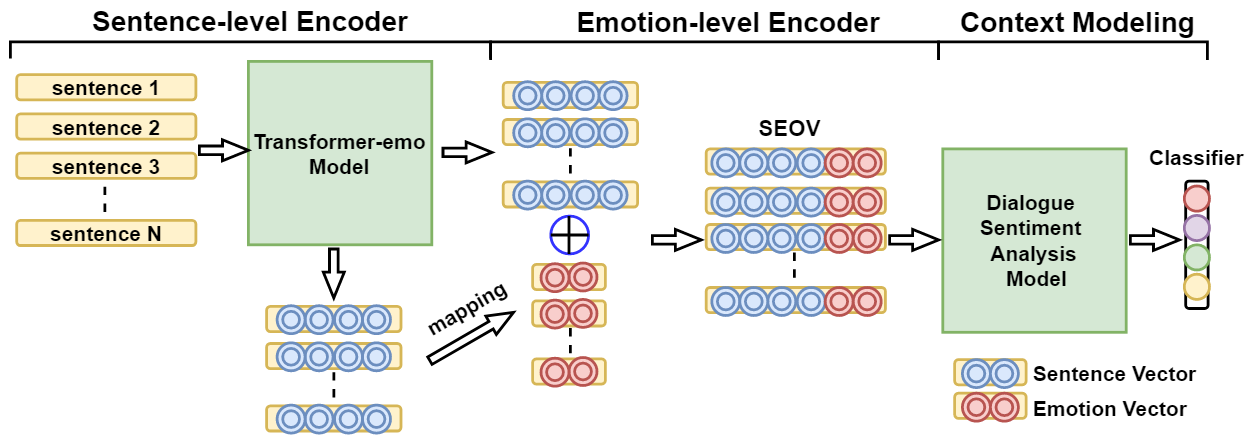}
	\caption{Our model's structure.} \label{fig3}
\end{figure}

\noindent \textbf{Sentence-level Encoder:}
Since the pre-trained language model can't directly encode the dialogue, we at first split the conversation $U_n$ into a series of single sentences :{$s_1, s_2, s_3...s_n$}. Then, we input them into an improved pre-trained language model called transformer-emo model, which can map the representation of utterance to the sentence vectors $Q$:   
\begin{equation}
	Q=Transformer-emo(s_1,s_2,s_3…s_n)
\end{equation}
where $Q$ is a series of sentence vectors :${q_1,q_2,q_3...q_n}$, the length of each sentence vector is set as $d$ ($d$ is set to 786 in our experiment).
The classical pre-trained language model BERT is proposed with transformer structure to train the general text corpus (Devlin et al., 2019). Compared with the CNN model (Kim,2014), transformer can obtain more syntactic and semantic information. In order to better adapting to our model, we improve the transformer model by adjusting the output form (referred to as transformer-emo) to obtain the sentence vectors of the utterance. 

\noindent \textbf{Emotion-level Encoder:}
As for sentence vectors, the length of the sentence itself, semantic information, symbols, etc. are the "size" of the vector, and its classification attributes are the "direction" of the vector. The sentence representation obtained by encoding the sentence in the existing method is not a "vector" in the true sense, because it only contains the "size" without the "direction". So we propose a new expression paradigm, SEOV, which represents the emotional intensity of sentence vector with its "size", and models the emotion tendency between emotion vectors with its "direction".

We map the vector $q$ from the k-dim space to the k*-dim space to obtain the emotion vector $q*$, to achieve emotions representation:
\begin{equation}
	q*=q[w_1,w_2,...w_{k^\ast}] 
\end{equation}
where $w_1,w_2,\ldots,w_{k^\ast}$ is the weight parameter.

In theory, when the spatial dimension after the mapping is equal to the number of categories, we can regard it as the classification result. The elements in the emotion vector $q*$ represent the probabilities of classified emotions in each sentence. 

Then we merge the original vector $q$ and the obtained emotion vector $q*$ to obtain the sentence-level emotion orientation vector e: 
\begin{equation}
	e=q\oplus q^\ast
\end{equation}     

The $e$ is called as SEOV, in which, the emotion’s direction is represented in the collection of SEOVs as shown in Fig.2. With the final DSAM, the “direction" of the emotion tendencies can be efficiently extracted and achieves better emotion recognition performance.

\noindent \textbf{Context Modeling:}
Finally, we reassemble SEOVs into dialogue lists, and input them into the DSAM. The existing DSAMs can distinguish the speaker and obtain the speaker state, so it can fully obtain the context information and context. We choose DialogueRNN, DialogueGCN, and bc-LSTM models as the benchmark version of the dialogue emotion analysis models to compare the test results.

\section{Experimental Setting}

\subsection{Datasets}

\textbf{IEMOCAP} (Busso et al. 2008): IEMOCAP dataset contains the conversation data of ten actors in the emotional interaction process, including video, voice, facial expression capture, and conversation text. The emotions are classified into six types of emotions: happy, sad, neutral, angry, excited, frustrated.

\noindent \textbf{MELD} (Poria et al. 2019): MELD dataset selects 1432 conversations from the TV series "Friends", with a total of 13,708 sentences, including video, text, voice and other data content. Emotions are classified into seven emotions: neutral, surprise, fear, sadness, joy, disgust, and angry.

% \usepackage{multirow}

\subsection{State-of-the-art Baselines}

\textbf{DialogueRNN} (Majumder et al. 2019): DialogueRNN uses different GRU units to obtain contextual information and speaker relationships. It is the first conversation sentiment analysis model to distinguish between speakers.

\noindent \textbf{DialogueGCN} (Ghosal et al. 2019): DialogueGCN constructs a conversation into a graph, transforms the speech emotion classification problem into a node classification problem of the graph, and uses the graph convolutional neural network to classify the results. 

\noindent \textbf{DialogXL} (Weizhou Shen et al. 2020): DialogXL use XLNet model for conversation emotion recognition to obtain longer-term contextual information.

\noindent \textbf{Bc-LSTM} (Poria et al. 2017): Bc-LSTM uses a two-way LSTM structure to obtain contextual semantic information, but does not distinguish between speaker states. 

\noindent \textbf{TRMSM} (Jiangnan Li et al. 2020): TRMSM uses the transformer structure to simplify the conversation relationship into Intra-Speaker and Inter-Speaker, which can solve the problem of long-distance context dependence.

\noindent \textbf{BERT} (Devlin et al. 2019): BERT is a pre-trained language model that can be fine-tuned to achieve good results in downstream tasks. In this article, we use BERT to classify the sentiment of a single sentence text and compare it with our model.

\section{Results and Analysis}

We compare our model with all baselines on the MELD and IEMOCAP datasets and get the experimental results in Tables 1. As expected, our proposal outperforms other baseline models.

% \usepackage{multirow}
\begin{table*}[]
	\caption{Experimental results(F1 score) on the IEMOCAP dataset and MELD dataset. SEOVER-RNN represents the result of using DialogueRNN as the fine-tuning model, SEOVER-GCN represents the result of using DialogueGCN as the fine-tuning model, and SEOVER-LSTM represents the result of using bc-LSTM as the fine-tuning model.}
	\begin{tabular}{c|c|c|c|c|c|c|c|c}
		\hline
		\multirow{2}{*}{Model} & \multicolumn{7}{c|}{IEMOCAP}                                                                                         & MELD           \\ \cline{2-9} 
		& Happy          & Sad            & Neutral        & Angry          & Excited        & Frustrated     & Average        & Average        \\ \hline
		DialogueRNN            & 33.18          & 78.80          & 59.21          & 65.28          & 71.86          & 58.91          & 62.75          & 55.90          \\
		DialogueGCN            & 42.75          & 84.54          & 63.54          & 64.19          & 63.08          & \textbf{66.99} & 64.18          & -              \\
		DialogXL               & -              & -              & -              & -              & -              & -              & 65.95          & 62.41          \\
		TRMSM                  & 50.22          & 75.82          & 64.15          & 60.97          & 72.70          & 63.45          & 65.74          & 62.36          \\
		bc-LSTM                & 43.40          & 69.82          & 55.84          & 61.80          & 59.33          & 60.20          & 59.19          & 55.90          \\
		BERT                   & -              & -              & ­-             & -              & -              & ­-             & 54.01          & 60.34          \\ \hline
		SEOVER-RNN             & 69.47          & 83.57          & \textbf{66.67} & 67.46          & \textbf{82.46} & 55.36          & 69.86          & \textbf{65.66} \\
		SEOVER-GCN             & 53.85          & 80.40          & 58.77          & 62.09          & 79.22          & 56.87          & 65.29          & -              \\
		SEOVER-LSTM            & \textbf{70.38} & \textbf{85.65} & 65.40          & \textbf{69.45} & 80.98          & 64.96          & \textbf{72.07} & 63.82          \\ \hline
	\end{tabular}
\end{table*}

\subsection{Comparison with baseline models}
Compared with the state-of-the-art conversation emotion recognition models, our proposal achieves better performance. This is cause by two reasons. Firstly, our model can obtain more syntactic and semantic features of utterances’ presentation by using transformer-emo which is a transformer based pretraining model. Secondly, the proposed SEOV can efficiently map the emotion orientations between sentence vectors. 

\subsection{Ablation Study}

In order to study the influence of the fusion of SEOV on the experimental results, we remove the emotion vectors in the SEOVER to compare the emotion recognition performance with benchmark model of DialogueRNN on the MELD dataset. As shown in Table 2, unsurprisingly, the accuracy and F1 score of the model without emotion vectors are much lower. The comparison illustrates the importance of the emotion tendency encoding in SEOV for the conversation emotion recognition.

\begin{table}
	\centering
	\caption{Results of ablation experiments, where accuracy and F1 score are both weighted results.}
		\begin{tabular}{l|c|c}
			\hline
			Model            & Accuracy       & F1 score       \\ \hline
			DialogueRNN      & 56.10          & 55.90          \\
			DialogueRNN-BERT & 58.59          & 59.40          \\
			SEOVER-RNN       & \textbf{65.33} & \textbf{65.66} \\ \hline
	\end{tabular}
	
\end{table}

\subsection{Error Analysis}

We analyze the confusion matrix on the MELD data set. As shown in Table 3, assuming "sad" and "joy" are opposite emotions, the count of the misjudge between "sad" an "joy" is relatively small. On the other hand, the "sad" and "angry" are a pair of adjacent emotions. The misjudge count of this pair adjacent is relatively high. It is not difficult to find that the fusion of emotion vectors is a double-edged sword, which improves the classification accuracy of flipped emotions, but slightly reduced the classification accuracy of adjacent emotions. Therefore, in future research, we will focus on improving the performance of the adjacent emotions classification.

\begin{table*}[] \centering
	\caption{Confusion matrix obtained on the test set using the MELD dataset and DialogueRNN as the benchmark model}	
	\begin{tabular}{cccccccc}
		\hline
		\multicolumn{8}{c}{Confusion Matrix}                                                    \\ \hline
		\multicolumn{1}{l|}{}         & neutral & surprise & fear & sad & joy & disgust & anger \\ \hline
		\multicolumn{1}{c|}{neutral}  & 3781    & 164      & 30   & 176 & 428 & 33      & 135   \\
		\multicolumn{1}{c|}{surprise} & 28      & 134      & 1    & 20  & 52  & 4       & 41    \\
		\multicolumn{1}{c|}{fear}     & 14      & 6        & 1    & 6   & 8   & 5       & 10    \\
		\multicolumn{1}{c|}{sad}      & 44      & 20       & 3    & 68  & 28  & 7       & 38    \\
		\multicolumn{1}{c|}{joy}      & 71      & 27       & 2    & 14  & 272 & 2       & 14    \\
		\multicolumn{1}{c|}{disgust}  & 17      & 6        & 4    & 12  & 9   & 9       & 11    \\
		\multicolumn{1}{c|}{anger}    & 55      & 36       & 5    & 29  & 30  & 18      & 165   \\ \hline
	\end{tabular}
\end{table*}
\section{Conclusion}
In this work, we propose a new paradigm of sentence-level emotion orientation vector(SEOV) to assist the emotion recognition, which solves the discourse representation information loss problem of conventional methods in the CER. Then, we designed a conversation emotion recognition model based on the SEOV called as SEOVER. It uses the Transformer-emo model to encode sentence vectors and emotion vectors containing emotion tendency information and fuses them as SEOV. Then, the SEOV is leveraged as input to active the final dialogue emotion analysis model to classify the speaker’s emotions. We conducted comparative experiments of the proposal with several benchmarks on both the MELD and IEMOCAP dataset. The experimental results prove that the proposed SEOVER outperforms the state-of-the-art methods.


\end{document}